\documentclass[final]{cvpr}

\usepackage{times}
\usepackage{epsfig}
\usepackage{graphicx}
\usepackage{amsmath}
\usepackage{amssymb}
\usepackage{bm}
\usepackage{nicefrac}
\usepackage[accsupp]{axessibility}  

\usepackage{booktabs} 

\usepackage[pagebackref=true,breaklinks=true,colorlinks,bookmarks=false]{hyperref}

\def\cvprPaperID{6633} 
\def\confYear{CVPR 2022}

\newcommand\blfootnote[1]{%
  \begingroup
  \renewcommand\thefootnote{}\footnote{#1}%
  \addtocounter{footnote}{-1}%
  \endgroup
}

\begin{document}

\title{The Implicit Values of A Good Hand Shake:\\Handheld Multi-Frame Neural Depth Refinement
}

\author{\hspace{-5pt}Ilya Chugunov$^{1\dagger}$\;\;  Yuxuan Zhang$^1$\;\; Zhihao Xia$^2$\;\; Xuaner Zhang$^2$\;\; Jiawen Chen$^2$\;\; Felix Heide$^1$ \vspace{5pt}\\
$^1$Princeton University \quad $^2$Adobe}

\maketitle
\blfootnote{$\dagger$ Developed data acquisition pipeline during Adobe internship.}

\begin{abstract}
    Modern smartphones can continuously stream multi-megapixel RGB images at 60~Hz, synchronized with high-quality 3D pose information and low-resolution LiDAR-driven depth estimates. During a snapshot photograph, the natural unsteadiness of the photographer's hands offers millimeter-scale variation in camera pose, which we can capture along with RGB and depth in a circular buffer. In this work we explore how, from a bundle of these measurements acquired during viewfinding, we can combine dense micro-baseline parallax cues with kilopixel LiDAR depth to distill a high-fidelity depth map. We take a test-time optimization approach and train a coordinate MLP to output photometrically and geometrically consistent depth estimates at the continuous coordinates along the path traced by the photographer's natural hand shake. With no additional hardware, artificial hand motion, or user interaction beyond the press of a button, our proposed method brings high-resolution depth estimates to point-and-shoot ``tabletop'' photography -- textured objects at close range.
\end{abstract}
\vspace{-1em}


\section{Introduction}
The cell-phone of the 90s was a phone, the modern cell-phone is a handheld computational imaging platform~\cite{delbracio2021mobile} that is capable of acquiring high-quality images, pose, and depth. Recent years have witnessed explosive advances in passive depth imaging, from single-image methods that leverage large data priors to predict structure directly from image features~\cite{ranftl2021vision,ranftl2019towards} to efficient multi-view approaches grounded in principles of 3D geometry and epipolar projection~\cite{tankovich2021hitnet, shamsafar2021mobilestereonet}. At the same time, progress has been made in the miniaturization and cost-reduction~\cite{callenberg2021low} of active depth systems such as LiDAR and correlation time-of-flight sensors~\cite{lange2001solid}. This has culminated in their leap from industrial and automotive applications~\cite{schwarz2010mapping,dong2017lidar} to the space of mobile phones. Nestled in the intersection of high-resolution imaging and miniaturized LiDAR we find modern smartphones, such as the iPhone 12 Pro, which offer access to high frame-rate, low-resolution depth and high-quality pose estimates.

\begin{figure}[t]
    \centering
    \includegraphics[width=\linewidth]{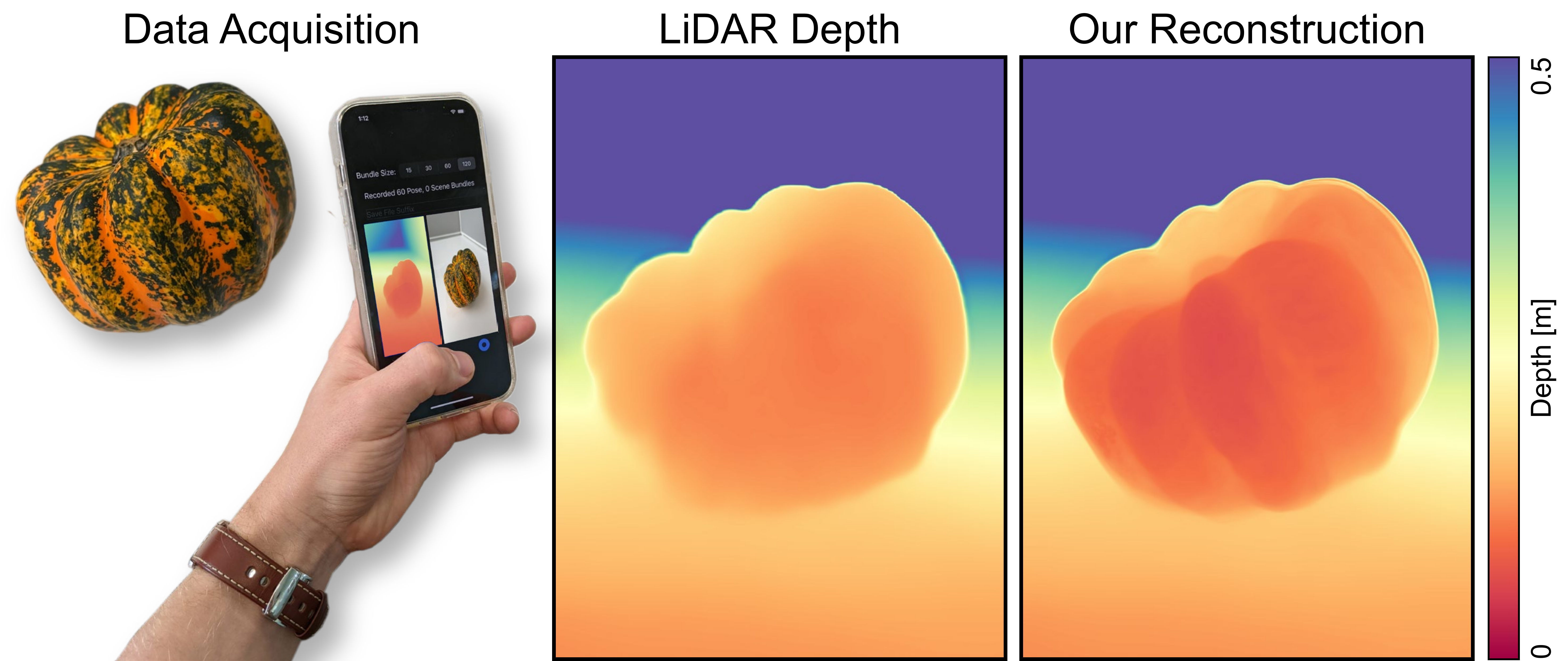}
    \caption{We reconstruct centimeter-scale depth features for this tabletop object from nothing more than a handheld snapshot.}
    \label{fig:my_label}
    \vspace{-1.5em}
\end{figure}

As applications of mixed reality grow, particularly in industry~\cite{li2018critical} and healthcare~\cite{gerup2020augmented} settings, so does the demand for convenient systems to extract 3D information from the world around us. Smartphones fit this niche well, as they boast a wide array of sensors -- e.g. cameras, magnetometer, accelerometer, and the aforementioned LiDAR system -- while remaining portable and affordable, and consequently ubiquitous. Image, pose, and depth data from mobile phones can drive novel problems in view synthesis~\cite{mildenhall2020nerf, park2021nerfies}, portrait relighting~\cite{pandey2021total,sun2019single}, and video interpolation~\cite{bao2019depth} that either implicitly or explicitly rely on depth cues, as well as more typical 3D understanding tasks concerning salient object detection~\cite{zhang2021bts,fan2020rethinking}, segmentation~\cite{schwarz2018rgb}, localization~\cite{zhuang2021semantic}, and mapping~\cite{schops2019bad, mur2017orb}.  

Although 3D scene information is essential for a wide array of 3D vision applications, today's mobile phones do not offer accurate high-resolution depth \emph{from a single snapshot}. While RGB image data is available at more than 100 megapixels (e.g. Samsung ISOCELL HP1), the most successful depth sensors capture at least three orders of magnitude fewer measurements, with pulsed time-of-flight sensors~\cite{morimoto2020megapixel} and modulated correlation time-of-flight imagers~\cite{lange20003d,hansard2012time,kolb2010time} offering kilopixel resolutions. Passive approaches can offer higher spatial resolution by exploiting RGB data; however, existing methods relying on stereo~\cite{chen20173d, Chang2018, Kendall2017} depth estimation require large baselines, monocular depth methods~\cite{chen2016monocular,ranftl2021vision} suffer from scale ambiguity, and structure-from-motion methods~\cite{schonberger2016structure} require diverse poses that are not present in a single snapshot. Accurate high-resolution snapshot depth remains an open challenge. 

For imaging tasks, \emph{align and merge} computational photography approaches have long exploited subtle motion cues \emph{during a single snapshot capture}. These take advantage of the photographer's natural hand tremor during viewfinding to capture a sequence of slightly misaligned images, which are fused into one super-resolved image~\cite{wronski2019handheld, tsai1984multiframe}. These misaligned frames can also be seen as mm-baseline stereo pairs, and works such as \cite{yu20143d, joshi2014micro} find that they contain enough parallax information to produce coarse depth estimates. Unfortunately, this micro-baseline depth is not enough to fuel mixed reality applications alone, as it lacks the ability to segment clear object borders or detect cm-scale depth features. In tandem with high-quality poses from phone-based SLAM~\cite{durrant2006simultaneous} and low-resolution LiDAR depth maps, however, we can use the high-resolution micro-baseline depth cues to guide the reconstruction of a refined high-resolution depth map. We develop a pipeline for recording LiDAR depth, image, and pose bundles at 60~Hz, with which we can conveniently record 120 frame bundles of measurements during a single snapshot event.

With this hand shake data in hand, we take a test-time optimization approach to distill a high-fidelity depth estimate from hand tremor measurements. Specifically, we learn an implicit neural representation of the scene from a bundle of measurements. Depth represented by a coordinate multilayer perceptron (MLP) allows us to query for depth at floating point coordinates, which matches our measurement model, as we effectively traverse a continuous path of camera coordinates during the movement of the photographer's hand. We can, during training, likewise conveniently incorporate parallax and LiDAR information as photometric and geometric loss terms, respectively. In this way we search for an accurate depth solution that is consistent with low-resolution LiDAR data, aggregates depth measurements across frames, and matches visual features between camera poses similar to a multi-view stereo approach. Specifically, we make the following contributions
\begin{itemize}
    \item A smartphone app with a point-and-shoot user interface for easily capturing synchronized RGB, LiDAR depth, and pose bundles in the field.\vspace{-0.25em}
    \item An implicit depth estimation approach that aggregates this data bundle into a single high-fidelity depth map.\vspace{-0.25em}
    \item Quantitative and qualitative evaluations showing that our depth estimation method outperforms existing single and multi-frame techniques.
\end{itemize}
The smartphone app, training code, experimental data, and trained models are available at \href{https://github.com/princeton-computational-imaging/HNDR}{github.com/princeton-computational-imaging/HNDR} .
\section{Related Work}
\noindent\textbf{Active Depth Imaging.}\hspace{0.1em}
Active depth methods prove a scene with a known illumination pattern and use the returned signal to reconstruct depth. Structured light approaches use this illumination to improve local image contrast~\cite{zhang2018high, scharstein2003high} and simplify the stereo-matching process. Time-of-Flight (ToF) technology instead uses the travel time of the light itself to measure distances. Indirect ToF achieves this through measuring the phase differences in the returned light~\cite{lange20003d}, whereas direct ToF methods time the departure and return of pulses of light via avalanche photodiodes~\cite{cova1996avalanche} or single-photon detectors (SPADs)~\cite{mccarthy2009long}. The low spatial-resolution depth stream we use in this work relis on a LiDAR direct ToF sensor. While its mobile implementation comes with the caveats of low-cost SPADs~\cite{callenberg2021low} and vertical-cavity surface-emitting lasers~\cite{warren2018low}, with limited spatial resolution and susceptibility to surface reflectance, this sensor can provide robust \emph{metric} depth estimates, without scale ambiguity, on even visually textureless surfaces. We use this LiDAR depth data to regularize our network's outputs, avoiding local minima solutions for low-texture regions.


\vspace{0.5em}\noindent\textbf{Multi-View Stereo.}\hspace{0.1em}
Multi-view stereo (MVS) algorithms are passive depth estimation methods that infer the 3D shape of a scene from a bundle of RGB views and, optionally, associated camera poses. COLMAP~\cite{schonberger2016structure} estimates both poses and sparse depths by matching visual features across frames. The apparent motion of each feature in image space is uniquely determined by its depth and camera pose. Thus there exists an important relationship that, for a noiseless system, \emph{any pixel movement (disparity) not caused by a change in pose must be caused by a change in depth}. While classical approaches typically formulate this as an explicit photometric cost optimization~\cite{sinha2007multi,furukawa2009accurate,galliani2015massively}, more recent learning-based approaches bend the definition of \emph{cost} with learned visual features~\cite{yao2018mvsnet,tankovich2021hitnet,lipson2021raft}, which aid in dense matching as they incorporate non-local information into otherwise textureless regions and are more robust to variations in lighting and noise that distort RGB values. In our setup, with little variation in lighting or appearance and free access to reliable LiDAR-based depth estimates in textureless regions, we look towards photometric MVS to extract parallax information from our images and poses.

\vspace{0.5em}\noindent\textbf{Monocular Depth Prediction.}\hspace{0.1em}
Single-image monocular approaches~\cite{ranftl2019towards,ranftl2021vision,godard2019digging} offer \emph{visually reasonable} depth maps, where foreground and background objects are clearly separated but may not be at a correct scale, with minimal data requirements -- just a single image.  Video-based methods such as \cite{fonder2021m4depth,luo2020consistent,watson2021temporal} leverage structure-from-motion cues~\cite{ullman1979interpretation} to extract additional information on scene scale and geometry. Works such as \cite{ha2016high,im2018accurate} use video data with \emph{small} (decimeter-scale) motion, and \cite{joshi2014micro,yu20143d} explore micro-baseline (mm-scale) motion. As the baseline decreases, the depth estimation problem gradually devolves from MVS to effectively single-image prediction. Our work resides in the micro-baseline domain, using only millimeters of baseline information, but leverages \emph{metric} LiDAR depth and pose information to bypass the noisy search for affine depth solutions -- the cycle of identifying and matching sparse image features -- that previous works were forced to contend with. 
\vspace{-1em}\section{Neural Micro-baseline Depth}
\noindent\textbf{Overview.}\hspace{0.1em} When capturing a ``snapshot photograph'' on a modern smartphone, the simple interface hides a significant amount of complexity. The photographer typically composes a shot with the assistance of an electronic viewfinder, holding steady before pressing the shutter. During composition, a modern smartphone streams the recent past, consisting of synchronized RGB, depth, and six degree of freedom pose (6DoF) frames into a circular buffer at 60~Hz.

In this setting, we make the following observations:
(1) A few seconds is sufficient for a typical snapshot of a static object.
(2) During composition, the amount of hand shake is small (mm-scale).
(3) Under small pose changes view-dependent lighting effects are minor.
(4) Our data shows that current commercial devices have excellent pose estimation, likely due to well-calibrated sensors (IMU, LiDAR, RGB camera) collaborating to solve a smooth low-dimensional problem.
Concretely, at each shutter press, we capture a ``data bundle'' of time-synchronized frames, each consisting of an RGB image $I$, 3D poses $P$, camera intrinsics $K$, and a depth map $Z$. In our experiments, the high-resolution RGB is 1920$\times$1440, while the LiDAR depth is 256 $\times$ 192\footnote{Though we refer to this as \emph{LiDAR depth}, the iPhone's depth stream appears to also rely on monocular depth cues. It unfortunately does not offer direct access to raw LiDAR measurements.}. To save memory, we restrict bundles to $N = 120$ frames (two seconds) in all our experiments.

\noindent\textbf{Micro-Baseline parallax.}\hspace{0.1em}
We specialize classical multi-view stereo~\cite{10.5555/861369} for our small motion scenario. Without loss of generality, we denote the first frame in our bundle the \emph{reference} (r) and represent all other \emph{query} (q) poses using the small angle approximation relative to the reference
\begin{equation}\label{eq:pose}
P_q = \left[\begin{array}{c|c} 
R(\bm{r}) & \bm{t}
\end{array}\right]
\approx  
\left[\begin{array}{ccc|c}
1 & -\bm{r}^{z} & \bm{r}^{y} & \bm{t}^{x} \\
\bm{r}^{z} & 1 & -\bm{r}^{x} & \bm{t}^{y} \\
-\bm{r}^{y} & \bm{r}^{x} & 1 & \bm{t}^{z} \\
\end{array}\right].
\end{equation}

Let $X$ the homogeneous coordinates of a 3D point $(x,y,z,1)^\top$, the \emph{geometric consistency} constraint is

\begin{equation}\label{eq:geometric_constraint}
X_r = P_q X_q.
\end{equation}
In other words, the known pose should transform any 3D point in the query frame to its corresponding 3D location in the reference  frame. Given camera intrinsics $K_r, K_q$, with
\begin{equation}\label{eq:intrinsics}
    K = 
\left[\begin{array}{ccc}
f_{x} & 0 & c_{x} \\
0 & f_{y} & c_{y} \\
0 & 0 & 1
\end{array}\right],
\end{equation}
and a point $X$ perspective projection yields continuous pixel coordinates $\bm{x}_r,\bm{x}_q$ via
\begin{equation}\label{eq:backward_model}
\bm{x}
=
\bm{\pi}(KX)
=
\left[\begin{array}{c}
u = \nicefrac{f_x x}{z} + c_x\\
v = \nicefrac{f_y y}{z} + c_y
\end{array}\right].
\end{equation}

\begin{figure}[t]
    \centering
    \includegraphics[width=\linewidth]{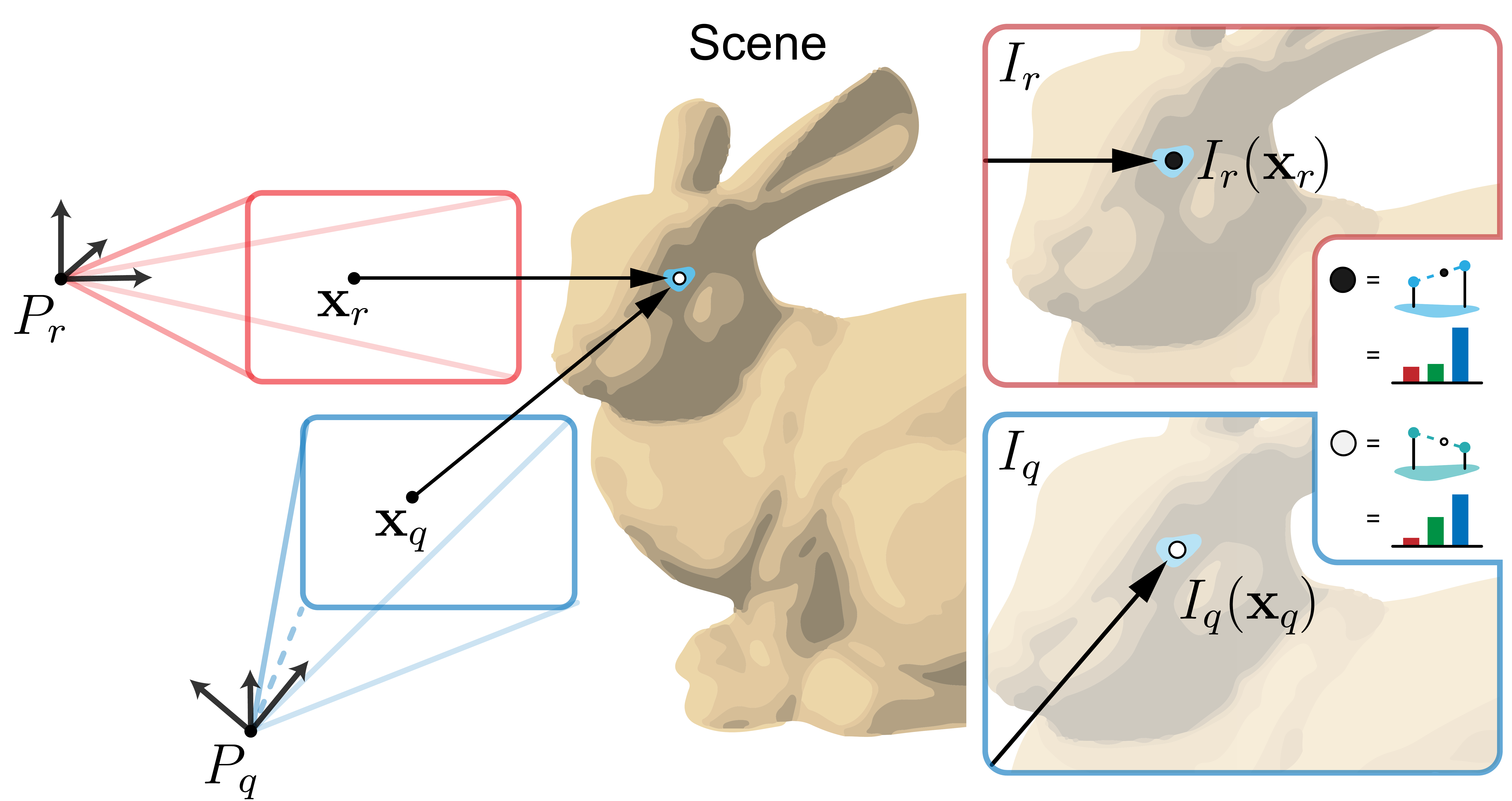}
    \caption{Visualization of how we project points $\bm{x}_r,\bm{x}_q$ with corresponding camera poses $P_r,P_q$ to 3D space and bilinearly sample image points $I_r(\bm{x}_r)$ and $I_q(\bm{x}_q)$. Note that the pose change here is enlarged for ease of illustration, the real misalignment in views from hand shake is on the scale of millimeters.}
    \label{fig:parallax_constraints}
    \vspace{-1em}
\end{figure}

\noindent Using these pixel coordinates to sample from images $I_r$ and $I_q$, we arrive at our second constraint
\begin{equation}\label{eq:photometric_constraint}
    I_r(\bm{x}_r = \bm{\pi}(K_r P_q X_q)) = I_q(\bm{x}_q).
\end{equation}
Corresponding 3D points should be \emph{photometrically} consistent: with small motion, they should have the same color in both views. These two constraints are visualized in Fig.~\ref{fig:parallax_constraints}. To generate a 3D point $X(\bm{x}, z)$ we can ``unproject'' a pixel coordinate $\bm{x}$ at depth $z = Z(\bm{x})$ by way of
\begin{equation}\label{eq:forward_model}
X(\bm{x}, z) = \bm{\pi}^{-1}(\bm{x},z;K)
=
\left[\begin{array}{c}
z(u-c_x) / f_x \\
z(v-c_y) / f_y \\
z \\
1
\end{array}\right].
\end{equation}
Our objective is to find a refined depth representation $Z'$ such that for all $z\in Z'$ any unprojected point $X(\bm{x})$ best satisfies our geometric~\eqref{eq:geometric_constraint} and photometric~\eqref{eq:photometric_constraint} constraints for all query views.


\begin{figure*}[t]
    \centering
    \includegraphics[width=\linewidth]{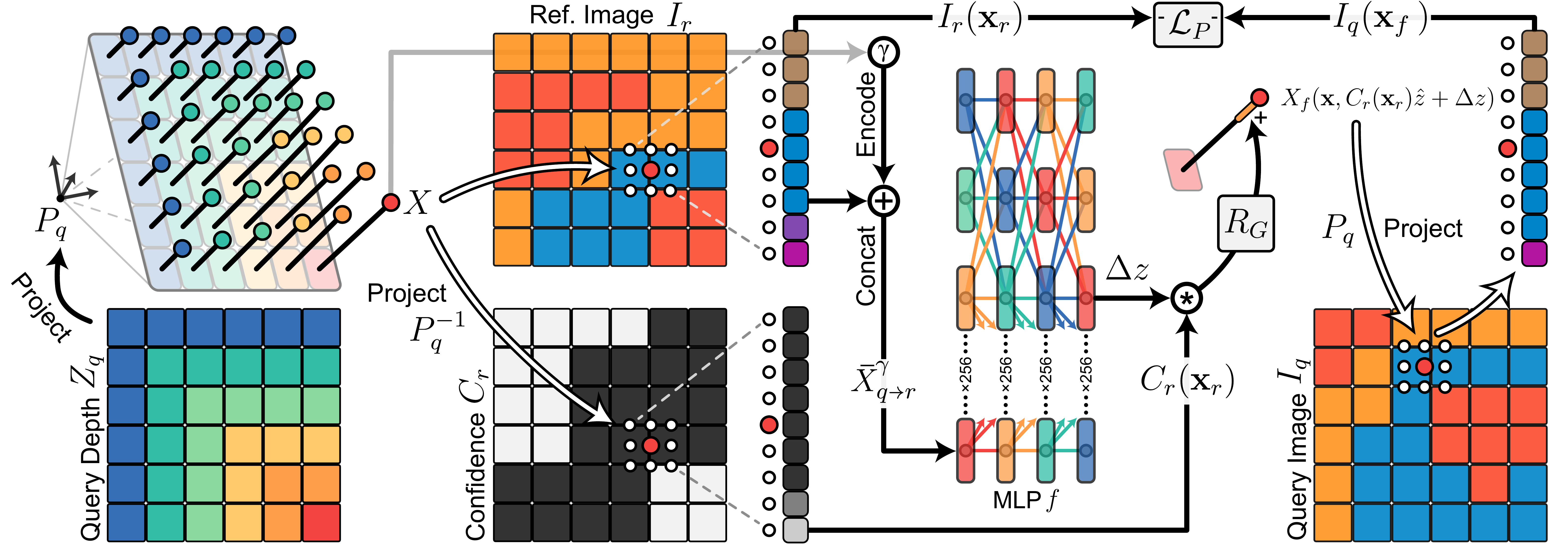}
    \caption{An illustrated pipeline of our proposed model. The query depth is used to project and sample a patch from our reference image $I_r$ for input into the MLP. This is weighed by a sample patch from our confidence $C_r$ to produce a depth offset $C_r(\bm{x_r})\Delta z$, which is used to project back to our query image $I_q$ and sample an image patch for loss calculation.}
    \label{fig:method}
    \vspace{-1em}
\end{figure*}

\vspace{0.5em}\noindent\textbf{Implicit Depth Representation.}\hspace{0.1em} There are numerous ways with which we can represent $Z'$. For example, we can represent it \emph{explicitly} with a discrete depth map from the reference view, or as a large 3D point cloud. While explicit representations have many advantages (fast data retrieval, existing processing tools), they are also challenging to optimize. Depth maps are discrete arrays and merging multiple views at continuous coordinates requires resampling. This blurs fine-scale information and is non-trivial at occlusion boundaries. Point cloud representations trade off this adaptive filtering problem with one of scale. Not only is a two second sequence with 120 million points unwieldy for conventional tools~\cite{arun1987least}, the points are almost entirely redundant.

Thus, we choose an \emph{implicit} depth representation in the form of a coordinate multi-layer perceptron~(MLP)~\cite{hornik1989multilayer}, where its learnable parameters automatically adapt to the scene structure.
Recent work have used MLPs to great success in neural rendering~\cite{mildenhall2020nerf,chen2021mvsnerf,park2021nerfies} and depth-estimation, where a continuous representation is of interest~\cite{zhang2021consistent}.
In our application, the MLP is a differentiable function
\begin{equation}
    z' = Z'(\bm{x}) = f(\mathrm{inputs}; \theta)
\end{equation}
returning a continuous $z'$ given an encoding of position, camera pose, color, and other features. In our implementation, $\mathrm{inputs}$ is a positionally encoded 3D \emph{colored point}
\begin{equation}
   \bar{X}^\gamma=[\gamma(x), \gamma(y), \gamma(z), r, g, b]^\top.
\end{equation}
We follow the encoding of \cite{mildenhall2020nerf} with
\begin{equation}\label{eq:positional_encoding}
    \gamma(p)=\left[\sin \left(2^{0} \pi p\right), \cos \left(2^{0} \pi p\right), \ldots ,\cos\left(2^{L-1} \pi p\right)\right],
\end{equation}
where $L$ is a selected number of encoding functions, and $r,g,b$ are color values scaled to $[0, 1]$. As the size of this MLP is fixed, the large dimensionality of our measurements does not affect the calculation of $z'$, and instead becomes a large training dataset from which to sample. Translating \eqref{eq:geometric_constraint} and \eqref{eq:photometric_constraint} into a regularized loss function on $z'$, and backpropagating through $f$, our implicit depth representation $\theta$ can be learned with stochastic gradient descent. 

\vspace{0.5em}\noindent\textbf{Backward-Forward Projection Model.}\hspace{0.1em} Fig. \ref{fig:method} illustrates how we combine geometric and photometric constraints to optimize our MLP to produce a refined depth. At each training step we sample a query view $(I_q, P_q, Z_q)$ and generate $M$ randomly sampled colored points $\bar{X}_q$ via Eq.~\ref{eq:forward_model}
\begin{align}\label{eq:ray_generation}
    &\bar{X}_q =
\left[\begin{array}{c}
\left[x,y,z,1\right]^\top \\
\left[r,g,b\right]^\top
\end{array}\right]
=
\left[\begin{array}{c}
X_q(\bm{x},z=Z_q(\bm{x})) \\
I_q(\bm{x})
\end{array}\right]
\nonumber
\\
&\bm{x} = \left[u,v\right]^\top, \quad u \sim \mathcal{U}(0,W), \quad v \sim \mathcal{U}(0,H),
\end{align}
where $H$ and $W$ are the image height and width, respectively. Here, $\bm{x}$ is a continuous coordinate and $I(\bm{x}), Z(\bm{x})$ represent sampling with a bilinear kernel. Following \eqref{eq:geometric_constraint} we transform these points to the reference frame as
\begin{equation}\label{eq:project_to_ref}
    \bar{X}_{q \rightarrow r} =
\left[\begin{array}{c}
\left[\hat{x},\hat{y},\hat{z},1\right]^\top \\
I_q(\bm{x})
\end{array}\right]
=
\left[\begin{array}{c}
P_q X_q(\bm{x},z) \\
I_q(\bm{x})
\end{array}\right]
.
\end{equation}

\noindent Then, rather than directly predicting a refined depth $z'$, we ask our MLP to predict a \emph{depth correction} $\Delta z$, that is
\begin{align}\label{eq:mlp_forward}
&\Delta z = f(\bar{X}^\gamma_{q \rightarrow r}) \\ \nonumber
&X_f(\bm{x},z')=X_f(\bm{x},\hat{z}+\Delta z) = \left[\hat{x},\hat{y},\hat{z} + \Delta z,1\right]^\top. \nonumber
\end{align}

\noindent As we show in Section \ref{sec:results}, this parameterization allows us to avoid local minima in poorly textured regions. We transform these refined points $X_f$ back to the query frame and resample the query image at the updated coordinates
\begin{align}\label{eq:resample}
I_q(\bm{x}_f) = I_q(\bm{\pi}(P_q^{-1}X_f(\bm{x},\hat{z}+\Delta z))).
\end{align}


\noindent Finally, our photometric loss is
\begin{align}\label{eq:photometric_loss}
    &\mathcal{L}_P = |I_q(\bm{x}_f) - I_r(\bm{x}_r)|^2  \nonumber\\
    &\bm{x}_r = \bm{\pi}(K_r P_q X_q(\bm{x},z)),
\end{align}
which attempts to satisfy \eqref{eq:photometric_constraint} by encouraging the colors of the refined 3D points to be the same in both the query and reference frames. While \eqref{eq:photometric_constraint} works well in well-textured areas, it is fundamentally underconstrained in flat regions. Therefore, we augment our loss with a weighted geometric regularization term based on \eqref{eq:geometric_constraint} that pushes the solution towards an interpolation of LiDAR depth in these regions
\begin{equation}\label{eq:geometric_loss}
    \mathcal{R}_G = |X_f(\bm{x},\hat{z}+\Delta z) - P_q X_q(\bm{x},z))| \approx |\Delta z|.
\end{equation}
Our final loss is a weighted combination of these two terms
\begin{equation}\label{eq:loss}
    \mathcal{L} = \mathcal{L}_P + \alpha\mathcal{R}_G,
\end{equation}
where by tuning $\alpha$ we adjust how strongly our reconstruction adheres to the LiDAR depth initialization.
\begin{figure}[t!]
    \centering
    \includegraphics[width=0.99\linewidth]{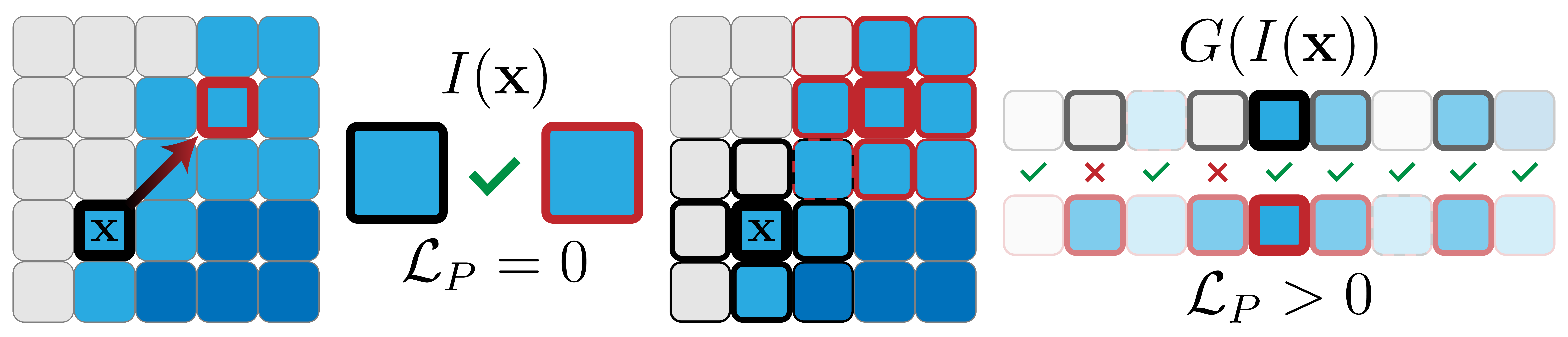}
    \caption{A weighted sampling around a point $\bm{x}$ we can avoid false matches in otherwise color-ambiguous image regions. Opacity and border thickness represents the weight of each sample.}
    \label{fig:gaussian-blob}
    \vspace{-1em}
\end{figure}

\vspace{0.5em}\noindent\textbf{Patch Sampling.}\hspace{0.1em}  In practice, we cannot rely on single-pixel samples as written in \eqref{eq:forward_model} for photometric optimization. Our two megapixel input images $I$ will almost certainly contain color patches that are larger than the depth-induced motion of pixels within them. With single-pixel samples, there are many incorrect depth solutions that yield $\mathcal{L}_P$ of zero. To combat this, we replace each sample $I(\bm{x})$ in \eqref{eq:ray_generation} with Gaussian-weighted patches
\begin{align}\label{eq:gaussian_blob}
    &G(I(\bm{x})) = \left[\mathcal{N}(\sqrt{\delta_u^2 + \delta_v^2};\mu,\sigma^2)I(\bm{x} - [\delta_u, \delta_v]^\top)\right], \nonumber\\
    &\mathrm{for} \quad \delta_u = \left\{-K \ldots K\right\}, \;\delta_v = \left\{-K \ldots K\right\}.
\end{align}
Fig.~\ref{fig:gaussian-blob} illustrates this for $K=3$: the increased receptive field discourages false color matches. Adjusting $K$, we trade off the ability to reconstruct fine features for robustness to noise and low-contrast textures (see supplement).

\vspace{0.5em}\noindent\textbf{Explicit Confidence.}\hspace{0.1em}  Another augmentation we make is to introduce a learned explicit $H\times W$ confidence map $C_r$ to weigh the MLP outputs. That is, we replace $\Delta z$ with $C_r(\bm{x}_r)\Delta z$ in \eqref{eq:mlp_forward}. This additional degree of freedom allows the network push $\Delta z$ toward zero in color-ambiguous regions, rather than forcing it to first learn a positional mapping of where these regions are located in the image. As $C_r(\bm{x}_r)$ only adds an additional sampling step during point generation, the overhead is minimal. Once per epoch we apply an optional $5\times5$ median filter to $C_r$ to minimize the effects of sampling noise during training.

\vspace{0.5em}\noindent\textbf{Final reconstruction.}\hspace{0.1em}  After training, to recover a refined depth map $Z^*$ we begin by reprojecting all low-resolution depth maps $Z_q$ to the reference frame following \eqref{eq:geometric_constraint}. We then average and bilinearly resample this data to produce a $H\times W$ depth map $Z_{avg}$. We query the MLP at $H\times W$ grid-spaced points, using $I_r$ and $Z_{avg}$ to generate points as in \eqref{eq:ray_generation}. Finally, we extract and re-grid the depth channel from the MLP outputs $R_f$ to produce $Z^*$.

\begin{figure}[t!]
    \centering
    \includegraphics[width=0.9\linewidth]{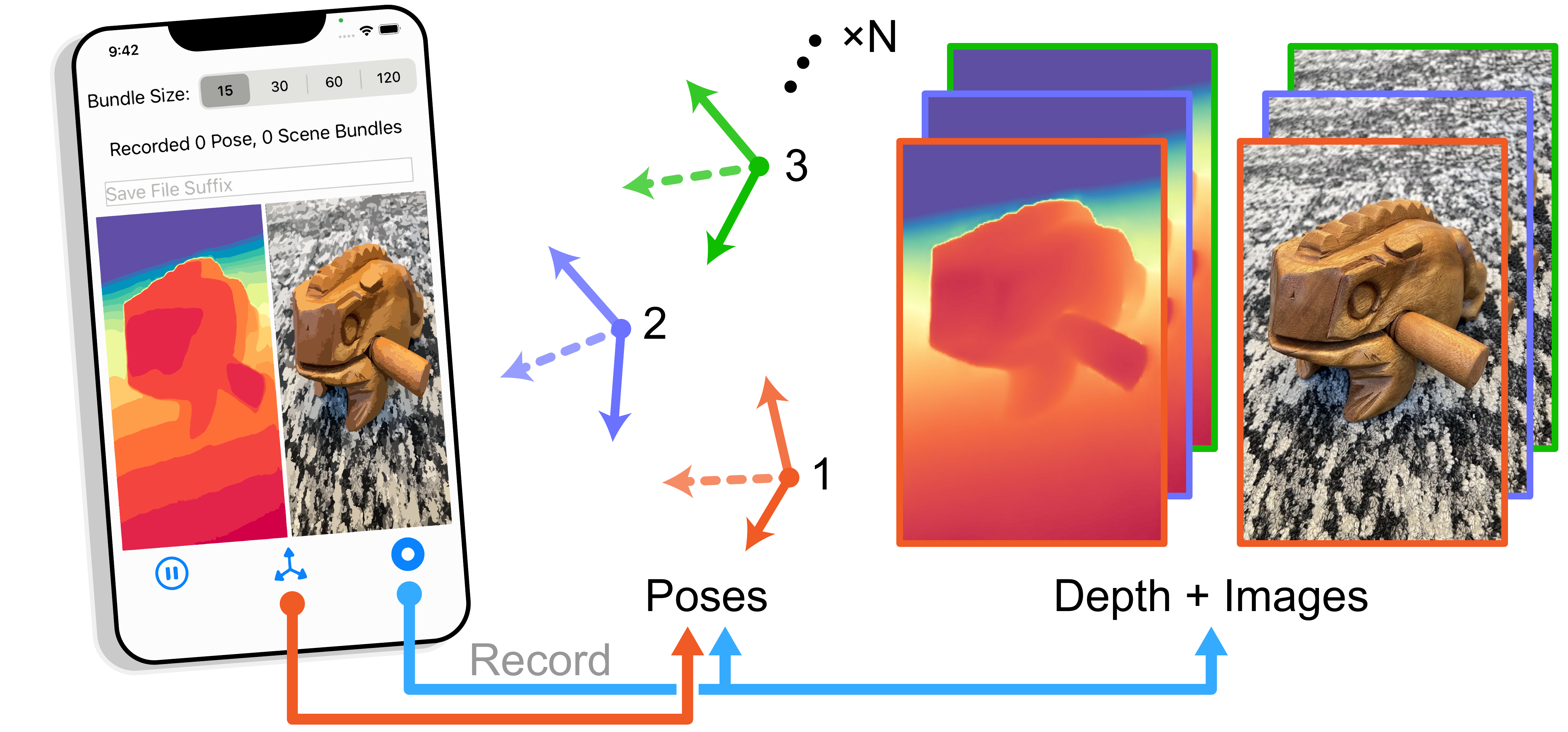}
    \caption{Our smartphone app for capturing data bundles, with two illustrated recording functions. The button to record only pose allows the user to save and analyze hundreds of hand motion bundles without the overhead of transferring tens of gigabytes of video.}
    \label{fig:data-collection}
    \vspace{-1em}
\end{figure}

\section{Data Collection}
\vspace{0.5em}\noindent\textbf{Recording a Bundle.}\hspace{0.1em} We built a smartphone application for recording bundles of synchronized image, pose, and depth maps. Our app, running on an iPhone 12 Pro using ARKit 5, provides a real-time viewfinder with previews of RGB and depth (Fig. \ref{fig:data-collection}). The user can select bundle sizes of [15, 30, 60, 120] frames ([0.25, 0.5, 1, 2] seconds of recording time) and we save all data, including nanosecond-precision timestamps to disk. We will publish the code for both the app and our offline processing pipeline (which does color-space conversion, coordinate space transforms, etc.).
\begin{figure}[h]
    \centering
    \includegraphics[width=\linewidth]{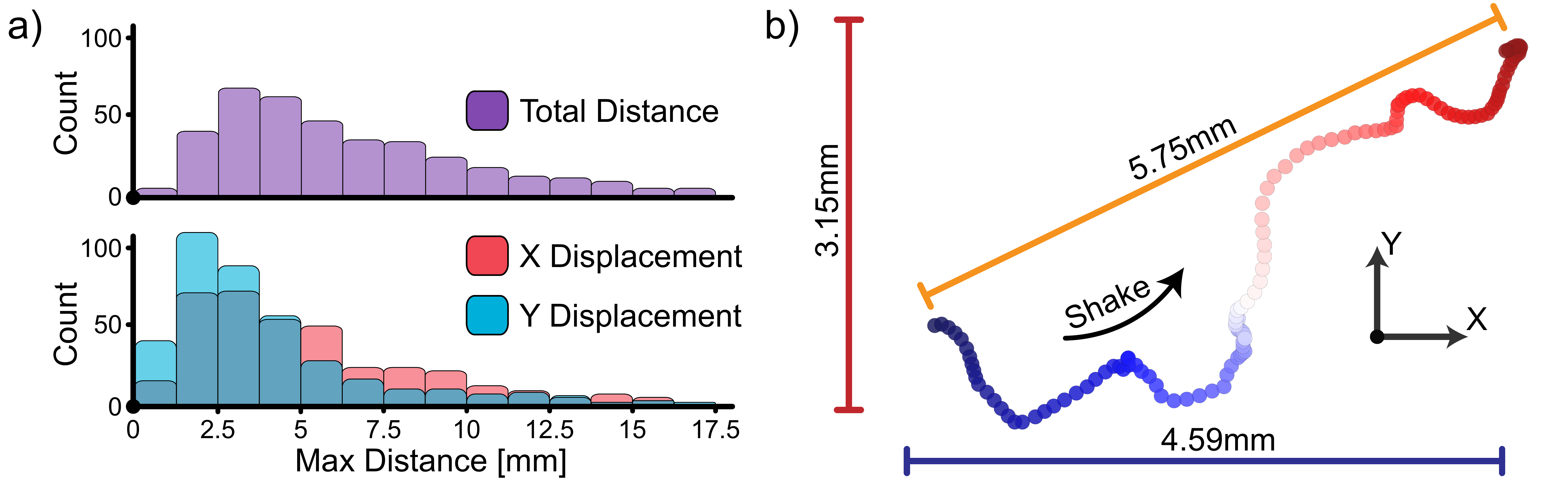}
    \caption{(a) Distribution of displacements in camera position during the capture of a 120 frame bundle in portrait mode. (b) Visualization of a hand tremor path with labeled median displacements.}
    \label{fig:hand-shake}
    \vspace{-1em}
\end{figure}

\vspace{0.5em}\noindent\textbf{Natural Hand Tremor Analysis.}\hspace{0.1em} To analyze hand motion during composition, we collected fifty 2-second pose-only bundles from 10 volunteers. Each was instructed to act as if they were capturing photos of objects around them, to hold the phone naturally in their dominant hand, and to keep focus on an object in the viewfinder. We illustrate our aggregate findings in Fig. \ref{fig:hand-shake} and individual measurements in the supplemental material. We focus on in-plane displacement they are the dominant contribution to observed parallax. We find that natural hand tremor appears similar to paths traced by 2D Brownian motion, with some paths traveling far from the initial camera position as in Fig.~\ref{fig:hand-shake} (b), and others forming circles around the initial position. Consequently, while the median effective baseline from a two second sequence is just under 6mm, some recordings exhibit nearly 1cm of displacement, while others appear are almost static. We suspect that the effects of breathing and involuntary muscle twitches are greatly responsible for this variance. Herein lies the definition of \emph{good} hand shake: it is one that produces a useful micro-baseline.
\begin{figure}[t!]
    \centering
    \includegraphics[width=\linewidth]{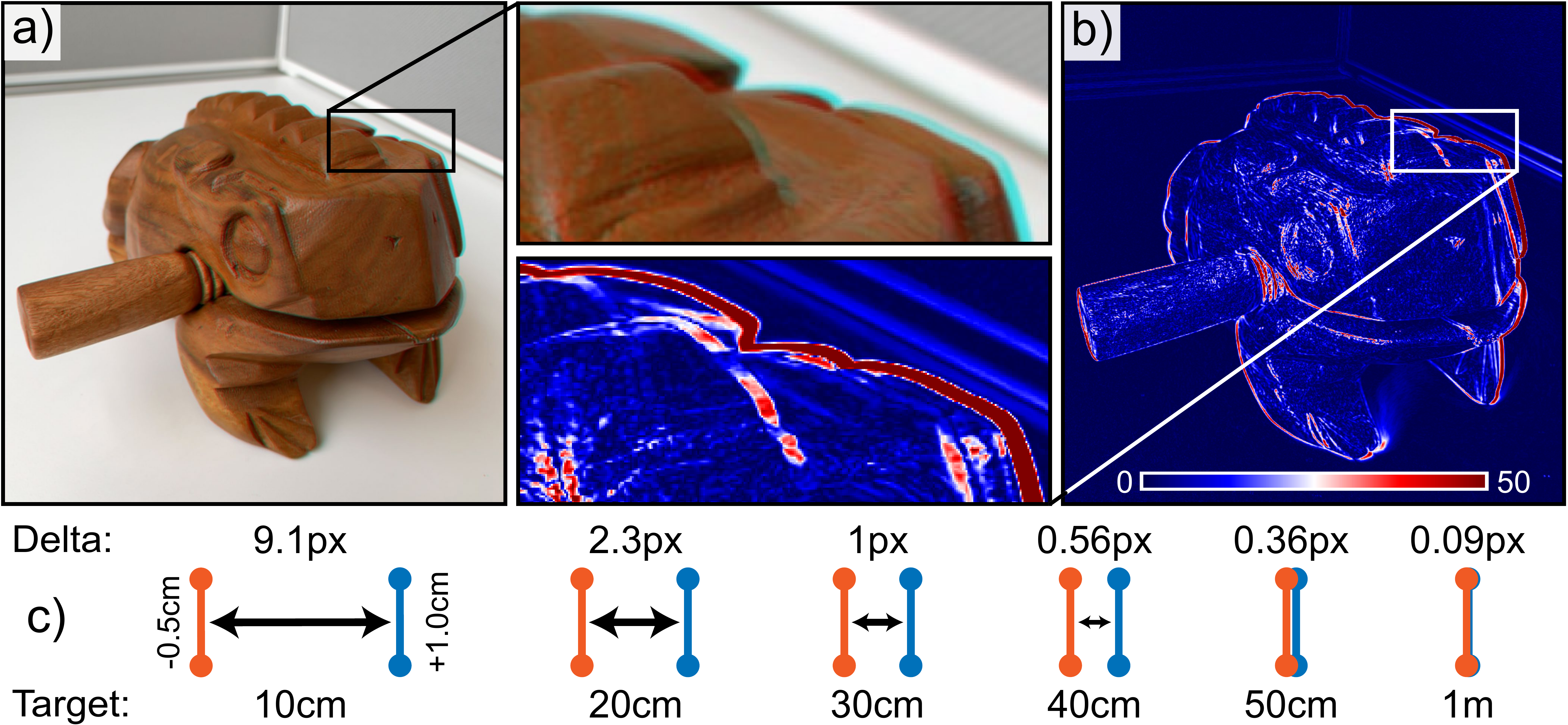}
    \caption{(a) Anaglyph visualization of maximum observed disparity for a 1m plane-rectified 120 frame image sequence. (b) Absolute difference between the frames in (a). (c) Observed disparity (Delta) with a 6mm baseline for a 1cm feature at depth (Target). }
    \label{fig:disparity-curve}
    \vspace{-1em}
\end{figure}
Fig. \ref{fig:disparity-curve} (c) illustrates, given our smartphone's optics, what a 6mm baseline translates to in pixel disparity and therefore the depth feature precision. We intentionally limit ourselves to a depth range of approximately \textbf{50~cm}, beyond which image noise and errors in pose estimation overpower our ability to estimate subpixel displacement.
\vspace{-0.5em}\section{Assessment}\label{sec:results}
\noindent\textbf{Implementation Details.}\hspace{0.1em} We use $N=120$ frame bundles for our main experiments. Images are recorded in portrait orientation, with $H=1920$, $W=1440$. Our MLP is a 4 layer fully connected network with ReLU activations and a hidden layer size of 256. For training we use inputs of $M=4096$ colored points, a kernel size of $K=11$, and $L=6$ encoding functions. We use the Adam optimizer~\cite{kingma2014adam} with an initial learning rate of $10^{-5}$, exponentially decreased over $200$ epochs with a decay rate of $0.985$. We apply the geometric regularization $\mathcal{R}_G$ with weight $\alpha=0.01$. We provide ablation experiments on the effects of many of these parameters in the supplement. Training takes about 45 minutes on an NVIDIA Tesla P100 GPU, or 180 minutes on an Intel Xeon Gold 6148 CPU.

\begin{figure*}[t]
    \centering
    \includegraphics[width=0.98\linewidth]{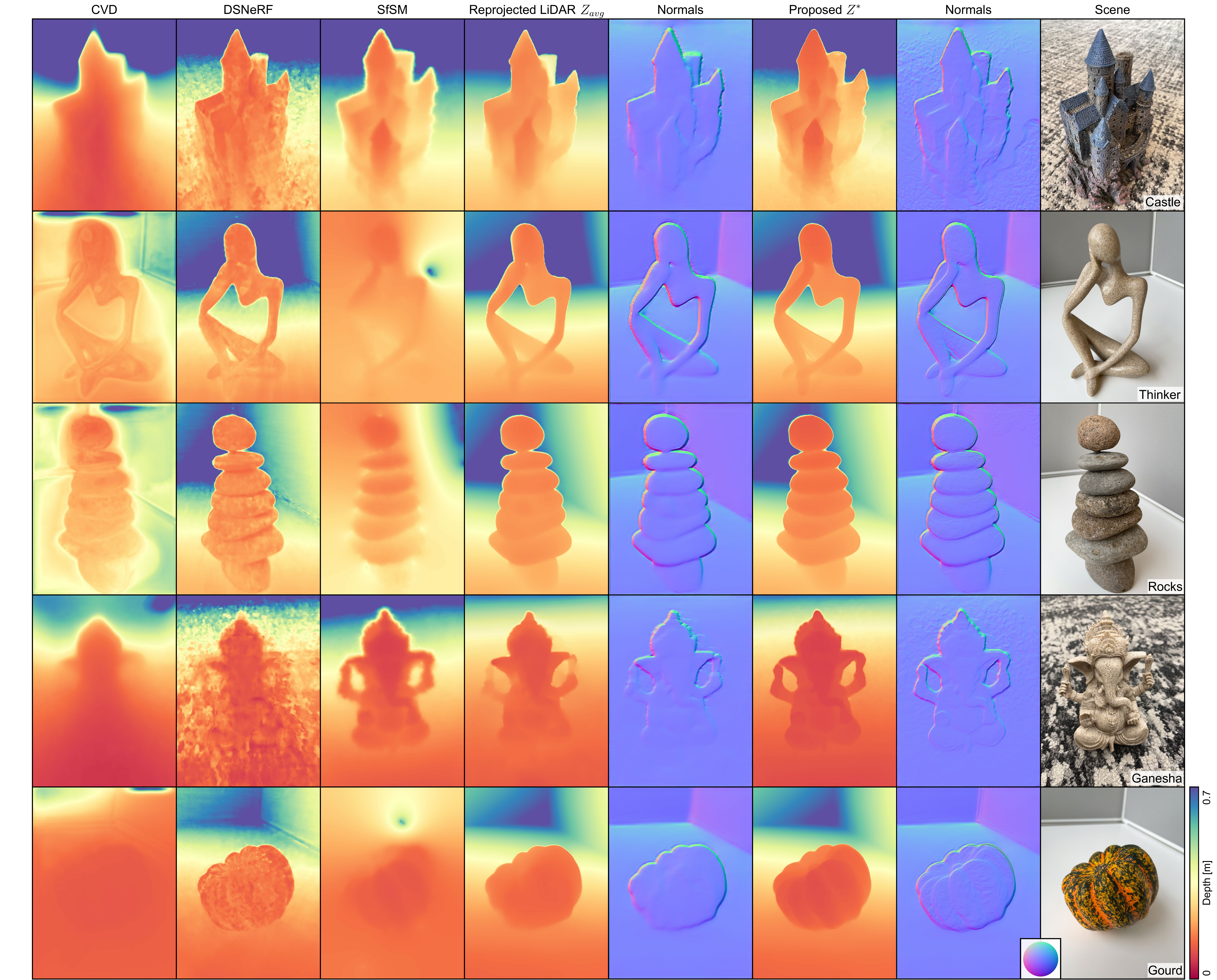}
    \caption{Qualitative comparison of depth reconstruction methods for tabletop scenes. Normals are shown for $Z_{avg}$ and our proposed method to highlight how we recover centimeter-scale features absent from the input depths. See the supplement for additional results.}
    \label{fig:results}
    \vspace{-1em}
\end{figure*}

\vspace{0.5em}\noindent\textbf{Comparisons.}\hspace{0.1em} We compare our method (\emph{Proposed}) to the input LiDAR data and several recent baselines. Namely, we reproject all the depth maps in our bundle to the reference frame and resample them to produce a $1920\times1440$ LiDAR baseline $Z_{avg}$. For the depth reconstruction methods we look to Consistent Video Depth Estimation~\cite{luo2020consistent} (CVD), which similarly uses photometric loss between frames in a video to refine a consistent depth; Depth Supervised NeRF~\cite{deng2021depth} (DSNeRF), which also features an MLP for depth prediction; and Structure from Small Motion~\cite{ha2016high} (SfSM), which investigates a closed-form solution to depth estimation from micro-baseline stereo. Both DSNeRF and CVD rely on COLMAP~\cite{schonberger2016structure} for poses or depth inputs. However, when input our micro-baseline data, COLMAP \emph{fails to converge} and returns neither. For fair comparison, we substitute our high-accuracy LiDAR poses and depths.

\begin{table}[t]
	\resizebox{\columnwidth}{!}{
	\begin{tabular}{ c c  c  c  c  c}
		\toprule
		\textbf{Scene} & CVD~\cite{luo2020consistent} & DSNeRF~\cite{deng2021depth} & SfSM~\cite{ha2016high} & $Z_{avg}$ & Proposed \\
			\midrule
			\midrule
		\emph{castle}      &4.88$/$62.9 &4.89$/$87.3 &4.70$/$55.4 &4.51$/$48.1 &\textbf{3.83$/$30.0} \\ 
		\emph{thinker}     &4.73$/$115&4.59$/$109 &5.78$/$170 &4.50$/$107 &\textbf{4.26$/$89.4} \\ 
		\emph{rocks}       &6.68$/$212 &5.83$/$180 &6.43$/$190 &5.52$/$163 &\textbf{4.59$/$97.3} \\ 
		\emph{ganesha}     &4.93$/$56.5 &5.29$/$103 &4.45$/$41.3  &4.41$/$36.8 &\textbf{3.51$/$23.4} \\ 
		\emph{gourd}       &4.25$/$92.6&4.00$/$108 &3.72$/$70.7&3.77$/$92.8&\textbf{3.14$/$52.9} \\ 
		\emph{eagle}       &3.93$/$63.1&4.09$/$93.4&6.60$/$219 &3.98$/$93.2&\textbf{3.32$/$49.0} \\ 
		\emph{double}       &4.77$/$68.6 &4.87$/$91.3 &4.30$/$39.1  &4.00$/$31.6&\textbf{3.57$/$22.1} \\ 
		\emph{elephant}    &5.12$/$101 &5.20$/$124 &5.49$/$120 &5.09$/$127 &\textbf{4.46$/$78.8} \\ 
		\emph{embrace}    &4.92$/$54.5 &5.13$/$88.7 &3.81$/$26.0 &3.27$/$23.1&\textbf{2.85$/$15.5} \\ 
		\emph{frog}        &3.44$/$68.2 &3.11$/$58.9&3.05$/$51.5&2.80$/$49.9&\textbf{2.58$/$34.1} \\ 
			\bottomrule
	\end{tabular}
	}
	\vspace{0.1em}
	\caption{\label{tab:photo_error}%
	Quantitative comparison of photometric error for our ten tested scenes. Each entry shows: mean absolute error $/$ mean squared error. Note that different scenes can have different scales of error, as it is dependent on their overall image texture content.
	}
\end{table}

\vspace{0.5em}\noindent\textbf{Experimental Results.}\hspace{0.1em} We present our results visually in Fig.~\ref{fig:results} and quantitatively in Table~\ref{tab:photo_error} in the form of photometric error (PE). To compute PE, we take the final depth map $Z^*$ output by each method, use the phone's poses and intrinsics to project each color point in $I_r$ to all other frames, and compare their sampled RGB values:
\begin{align}\label{eq:photometric_error}
    &PE = \Sigma|I_q(\bm{x}_q) - I_r(\bm{x})|, \quad \bm{x}_q=\bm{\pi}(P_q^{-1}X^*)\nonumber\\
    &X^{*} = \bm{\pi}^{-1}(\bm{x},Z^*(\bm{x});K), \quad \bm{x}=[u,v]^\top.
\end{align}
We exclude points $\bm{x}$ that transform outside the image bounds. Similar to traditional camera calibration or stereo methods, in the absence of ground truth depth, PE serves as a measure of how consistent our estimated depth is with the observed RGB parallax.
\begin{figure}[t]
    \centering
    \includegraphics[width=\linewidth]{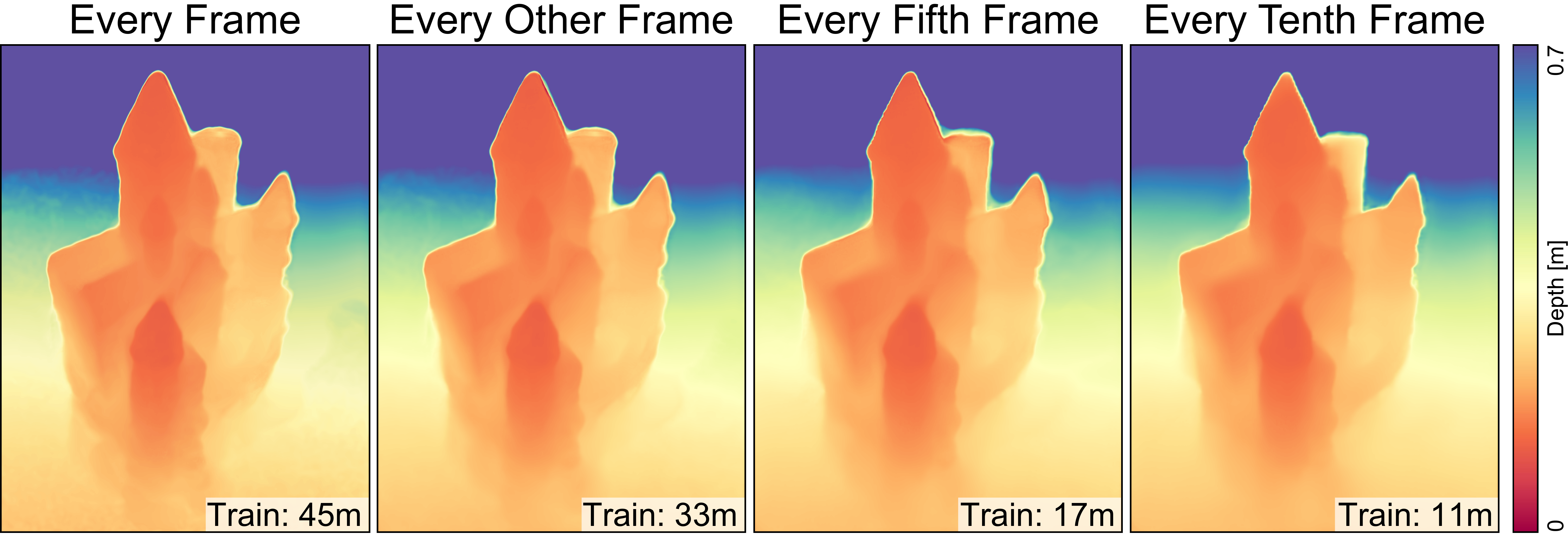}
    \caption{Frame count ablation with corresponding network train times. Note that due to various overheads, training time is not linear in the number of frames.}
    \label{fig:frame-count}
    \vspace{-1em}
\end{figure}
Table~\ref{tab:photo_error} summarizes the relative performance between these methods on 10 geometrically diverse scenes. Our method achieves the lowest PE for all scenes. Note that neither CVD nor DSNeRF achieve significantly lower PE as compared to the LiDAR depth $Z_{avg}$ even though both contain explicit photometric loss terms in their objective. We speculate that our micro-baseline data is out of distribution for these methods, and that the large loss gradients induced by small changes in pose results in unstable reconstructions. DSNeRF also has the added complexity of being a novel view synthesis method and is therefore encouraged to overfit to the scene RGB content in the presence of only small motion. We see this confirmed in Fig.~\ref{fig:results}, as DSNeRF produces an edge-aligned depth map but incorrect surface texture. SFsM successfully reconstructs textured regions close to the camera ($<$20cm), but fails for smaller disparity regions, textureless spaces, and parts of the image suffering from lens blur. The reprojected LiDAR depth produces well edge-aligned results but lacks intra-object structure, as it is relying on ambiguous mono-depth cues. Contrastingly, our proposed method reconstructs the \emph{castle}'s towers, the hand under the \emph{thinker}'s head, the depth disparity between stones in \emph{rocks}, the trunk and arms of \emph{ganesha}, and the smooth undulations of \emph{gourd}. For visually complex scenes such as \emph{ganesha} our method is able to cleanly separate the foreground object from its background.

\vspace{0.5em}\noindent\textbf{Frame Ablation.}\hspace{0.1em} Though the average max baseline in a recorded bundle is ~6mm, the baseline between neighboring frames is on the order of 0.1mm. This means that we need not use every frame for effective photometric refinement. This tradeoff between frame count, training time, and reconstruction detail is illustrated in Fig.~\ref{fig:frame-count}. We retain most of the \emph{castle} detail by skipping every other frame, and as we discard more data we see the reconstruction gradually lose fine edges and intra-object features. 

\begin{figure}[t]
    \centering
    \includegraphics[width=\linewidth]{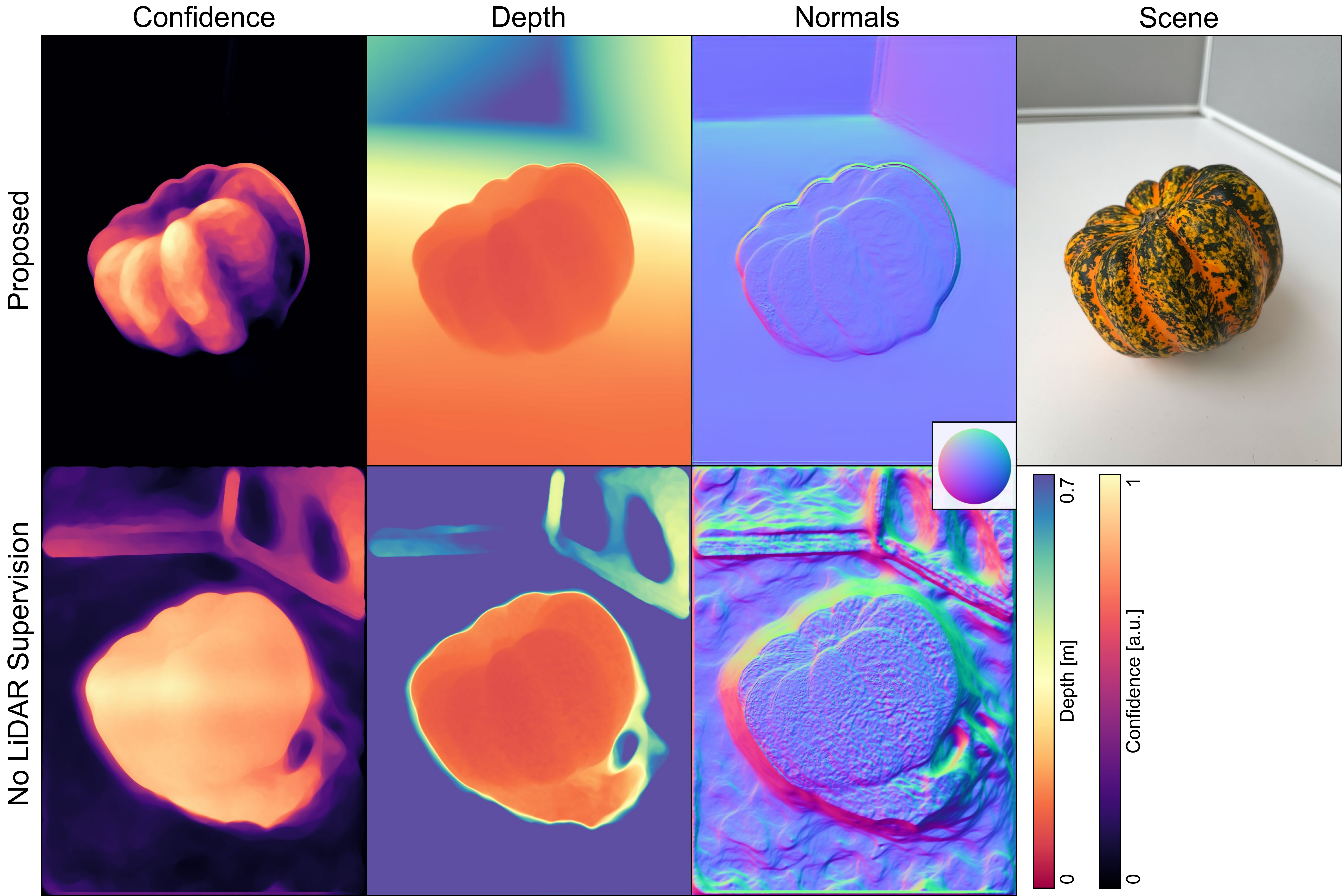}
    \caption{We reconstruct \emph{gourd} without LiDAR supervision to better understand the effects of geometric regularization.}
    \label{fig:photo-ablation}
    \vspace{-1em}
\end{figure}

\begin{figure}[t]
    \centering
    \includegraphics[width=\linewidth]{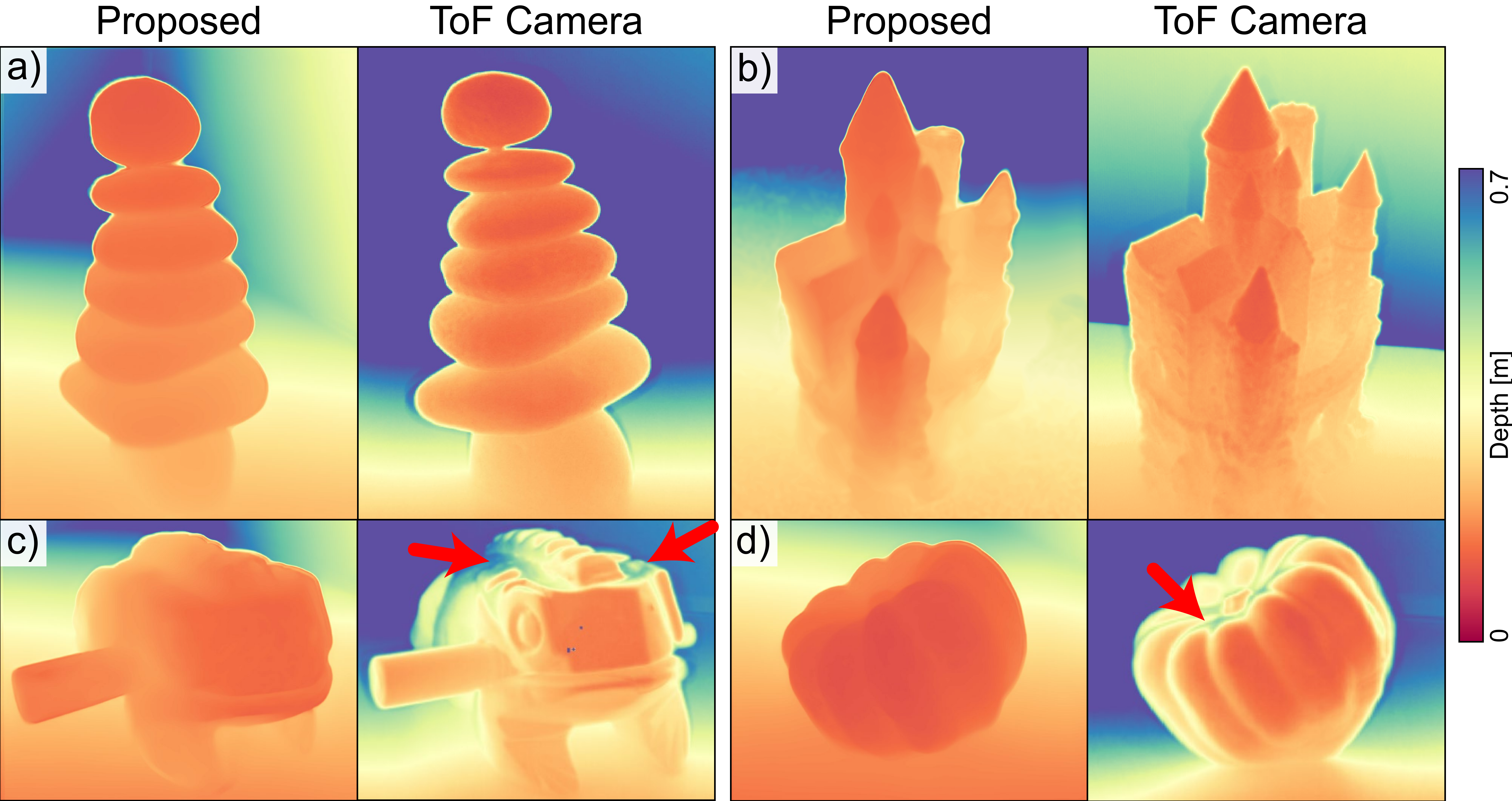}
    \caption{Qualitative comparison of proposed reconstruction results with depths captured by a time-of-flight camera. Examples (c) and (d) include arrows to highlight where the ToF camera suffers from severe multi-path interference artifacts and produces sharp depth discontinuities in place of expected smooth geometry.}
    \label{fig:tof-comparison}
    \vspace{-0.5em}
\end{figure}

\vspace{0.5em}\noindent\textbf{Role of LiDAR Supervision.}\hspace{0.1em} To determine the contribution of the low-resolution LiDAR data in our pipeline, we perform an ablation where we set $Z\;{=}\;$1m and disable our geometric regularizer by setting $\alpha\;{=}\;0$. Fig.~\ref{fig:photo-ablation} shows that, as expected, it fails to reconstruct the textureless background. It does, however, correctly reconstruct the gourd, and produces a result similar to the pipeline with full supervision. This demonstrates that our method extracts most of its depth details from the micro-baseline RGB images. LiDAR depth, however, is an effective regularizer: in regions without strong visual features, our learned confidence map $C_r$ is nearly zero and our reconstruction gracefully degrades to the LiDAR data. Finally, we note that even though this ablation discards depth supervision, the LiDAR sensor is still used by the phone to produce high-accuracy poses.

\vspace{0.5em}\noindent\textbf{Comparison to Dedicated ToF Camera.}\hspace{0.1em} We record several scenes with a high-resolution time-of-flight depth camera (LucidVision Helios Flex). Given the differences in optics and effective operating range, the viewpoints and metric depths do not perfectly match. We offset and crop (but don't rescale) the depth maps for qualitative comparison. Fig.~\ref{fig:tof-comparison} validates that our technique can reconstruct centimeter-scale features matching that of the ToF sensor, but smooths finer details corresponding to subpixel disparities. Since we rely on passive RGB rather than direct illumination, our technique can reconstruct regions the ToF camera cannot such as specular surfaces on the back of \emph{frog} and the top of \emph{gourd}. In these cases, multi-path interference leads to incorrect amplitude modulated ToF measurements.

\begin{figure}[t]
    \centering
    \includegraphics[width=\linewidth]{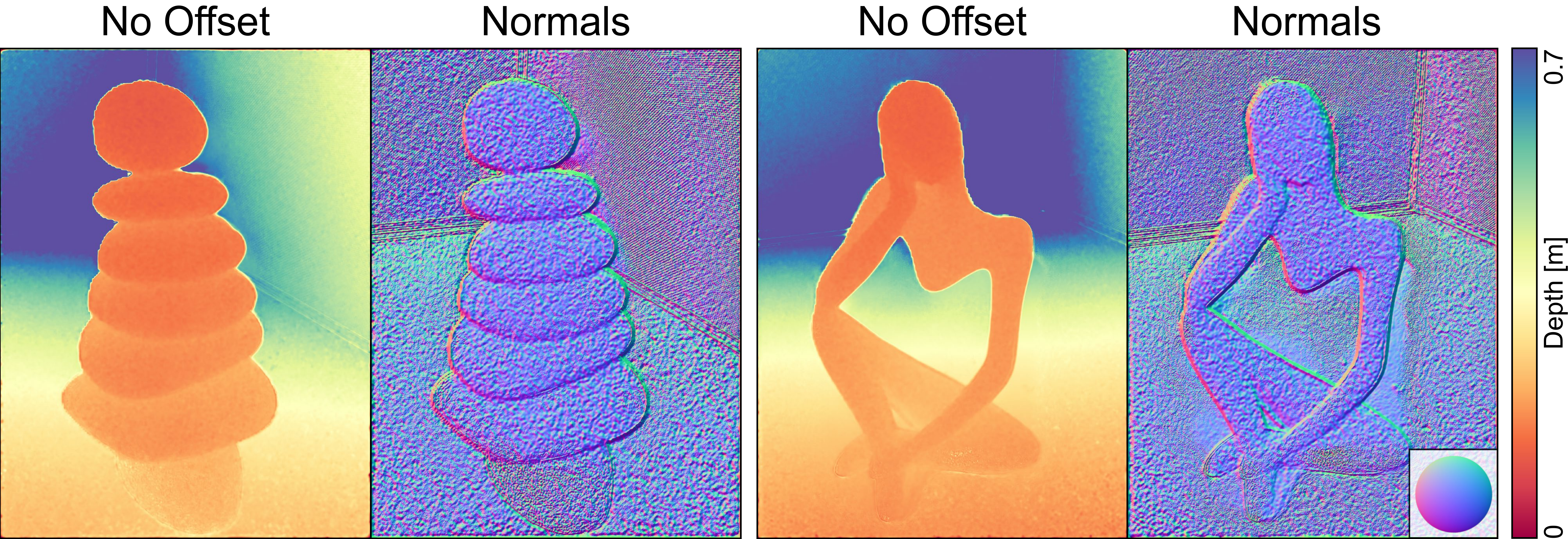}
    \caption{When we learn depth directly without a reasonable initialization we find that many samples end up stuck in local minima. This leads to noisy predictions where the MLP finds a false photometric matches far from the LiDAR depth estimate.}
    \label{fig:offset-ablation}
    \vspace{-1em}
\end{figure}

\noindent\textbf{Offsets over Direct Depth.}\hspace{0.1em} Rather than directly learn $Z^*$, we opt to learn offsets to the collected LiDAR depth data. We effectively start at a coarse, albeit smooth, depth estimate and for each location in space, and  search the local neighborhood for a more photometrically consistent depth solution. This allows us to avoid local minima solutions that overpower the regularization $\alpha R_G$ -- e.g. accidentally matching similar image patches. This proves essential for objects with repetitive textures, as demonstrated in Fig.~\ref{fig:offset-ablation}. 
\noindent Additional experiments can be found in the supplement.

\vspace{-0.5em}\section{Discussion and Future Work}
We show that with a modern smartphone, its possible to reconstruct a high-fidelity depth map from \emph{just a snapshot} of a textured ``tabletop'' object. We quantitatively validate that our technique outperforms several recent baselines and qualitatively compare to a dedicated depth camera. \\
\noindent\textbf{Rolling Shutter.} There is a delay in the tens of milliseconds between when we record the first and last row of pixels from the camera sensor~\cite{liang2008analysis}, during which time the position of the phone could slightly shift. Given accurate shutter timings, one may incorporate a model of rolling shutter similar to \cite{im2018accurate} directly into the implicit depth model. \\
\noindent\textbf{Training Time.} Although our training time is practical for offline processing and opens the potential for the easy collection of a large-scale training corpus, our method may be further accelerated with an adaptive sampling scheme which takes into account pose, color, and depth information to select the most useful samples for network training. \\
\noindent\textbf{Additional Sensors.} We hope in the future to get access to raw phone LiDAR samples, whose photon time tags could provide an additional sparse high-trust supervision signal. Modern phones now also come with multiple cameras with different focal properties. If synchronously acquired, their video streams could expand the overall effective baseline of our setup and provide additional geometric information for depth reconstruction -- towards snapshot smartphone depth imaging that exploits all available sensor modalities. \\
\noindent\textbf{Acknowledgments.}\hspace{0.1em} Ilya Chugunov was supported by an NSF Graduate Research Fellowship. Felix Heide was supported by an NSF CAREER Award (2047359), a Sony Young Faculty Award, and a Project X Innovation Award.

{\small
\bibliographystyle{ieee_fullname}
\bibliography{cvpr}
}

\end{document}